\documentclass[11pt,a4paper]{article}
\usepackage[hyperref]{acl2019}
\usepackage{times}
\usepackage{amsmath}
\usepackage{latexsym}

\usepackage{soul}
\usepackage{url}
\usepackage{multicol}
\usepackage[title]{appendix}
\usepackage{graphicx}
\usepackage{listings}
\newcommand{\COMMENT}[1] {}

\usepackage{url}

\aclfinalcopy 

\title{Training Neural Machine Translation To Apply Terminology Constraints}

\author{Georgiana Dinu  \hspace{0.8cm}
  Prashant Mathur \hspace{0.8cm}
  Marcello Federico \hspace{0.8cm}
  Yaser Al-Onaizan \\\\
  Amazon AI\\
  \small{\texttt{\{gddinu,pramathu,marcfede,onaizan\}@amazon.com}}
  }

\date{\today}

\begin{document}
\maketitle
\begin{abstract}

This paper proposes a novel method to inject custom terminology into neural machine translation at run time.  
Previous works have mainly proposed modifications to the decoding algorithm in order to constrain the output to include run-time-provided target terms. While being effective, 
these constrained decoding methods add, however, significant computational overhead to the inference step, and, as we show in this paper, can be brittle when tested in realistic conditions. In this paper we approach the problem by training a neural MT system to learn how to use custom terminology when provided with the input. Comparative experiments show that our method is not only more effective than a state-of-the-art implementation of constrained decoding, but is also as fast as constraint-free decoding. 
 
\end{abstract}

\section{Introduction}

Despite the high quality reached nowadays by neural machine translation (NMT), its output is often still not adequate for many specific domains handled daily by the translation industry. While NMT has shown to benefit from the availability of in-domain parallel or monolingual data to learn domain specific terms \cite{farajian2018evaluation}, it is not a universally applicable solution as often a domain may be too narrow and lacking in data for such bootstrapping techniques to work.
For this reason, most multilingual content providers maintain terminologies for all their domains which are created by language specialists. For example, an entry such as \textit{Jaws}~(en) $\rightarrow$ \textit{Lo Squalo}~(it) would exist in order to indicate that the input \textit{Jaws is a scary movie} should be translated as \textit{Lo Squalo \`e un film pauroso}. While  translation memories can be seen as ready-to-use training data for NMT domain adaptation, terminology databases (in short term bases) are more difficult to handle and there has been significant work on proposing methods to integrate domain terminology into NMT at run time.

Constrained decoding is the main approach to this problem. In short, it uses the target side of terminology entries whose source side match the input as decoding-time constraints. Constrained decoding and various improvements were addressed in \citet{chatterjee-EtAl:2017:WMT1}, \citet{Hasler2018},  \citet{HokampL17} among others. \citet{HokampL17} recently introduced the grid beam search (GBS) algorithm which uses a separate beam for each supplied lexical constraint. This solution however increases  the run time complexity of the decoding process exponentially in the number of constraints. \citet{PostV18} recently suggested using a dynamic beam allocation (DBA) technique that reduces the computational overhead to a constant factor, independent from the number of constraints.  
In practice,  results reported in \citet{PostV18}  show that  constrained decoding with DBA is effective but still causes a 3-fold increase in translation time when used with a beam size of 5.

In this paper we  address the problem of constrained decoding as that of learning a \emph{copy behaviour} of  terminology at \emph{training time}. By modifying the training procedure of neural MT we are completely eliminating any computational overhead at inference time. Specifically, the NMT model is trained to learn how to use terminology entries when they are provided as additional input to the source sentence. Term translations are inserted as inline annotations and additional input streams (so called source \textit{factors}) are used to signal the switch between running text and target terms. 
We present experiments on English-to-German translation with terms extracted from two terminology dictionaries. 
As we do not assume terminology is available at train-time, all our tests are performed in a \textbf{zero-shot} setting, 
that is with unseen terminology terms. We compare our approach against the efficient implementation of constrained decoding with DBA proposed by \citet{PostV18}.

While our goal resembles that of \citet{Gu2017} (teaching NMT to use translation memories) and of \citet{Pham2018} (exploring network architectures to enforce copy behaviour), the method we propose works with a standard transformer NMT model \cite{vaswani2017attention} which is fed a hybrid input containing running text and inline annotations. This decouples the terminology functionality from the NMT architecture, which is particularly important as the state-of-the-art architectures are continuously changing.

\section{Model}
We propose an integrated approach in which the MT model learns, at training time, how to use terminology when target terms are provided in input. In particular, the model should learn to \textit{bias} the translation to contain the provided terms, even if they were not observed in the training data. 
We augment the traditional MT input to contain a source sentence as well as a list of terminology entries that are triggered for that sentences, specifically those whose source sides match the sentence. 

\begin{table}[t]
\small
    \centering
    \begin{tabular}{l|l|p{50mm}}
        \hline
          Append & En  & All$_0$ \textbf{alternates}$_1$ \textbf{Stellvertreter}$_2$ shall$_0$ be$_0$  elected$_0$ for$_0$ one$_0$ term$_0$\\ 
         Replace & En & All$_0$ \textbf{Stellvertreter}$_2$ shall$_0$ be$_0$ elected$_0$ for$_0$ a$_0$ term$_0$\\ 
          \hline
          & De & Alle \textbf{Stellvertreter} werden f\"ur eine Amtszeit gew\"ahlt\\
         \hline
 
    \end{tabular}
    \caption{\small The two alternative ways used to generate source-target training data, including target terms in the source and factors indicating source words ($0$), source terms ($1$), and target terms ($2$). }
    \label{tab:annotation_example}
\end{table}
\noindent
While many different ways have been explored to augment MT input with additional information, we opt here for integrating terminology information as inline annotations in the source sentence, {by either appending the target term to its source version, or by directly replacing the original term with the target one.} We add an additional parallel stream to signal this ``code-switching" in the source sentence. When the translation is appended this stream has {three} possible values: 0 for source words (default), 1 for source terms, and 2 for target terms. The two tested variants, one in which the source side of the terminology is retained and one in which it is discarded, are illustrated with an example in Table \ref{tab:annotation_example}.

\subsection{Training data creation}
As we do not modify the original sequence-to-sequence NMT architecture, the network can learn the use of terminology from the augmentation of the training data.
We hypothesize that the model will learn to use the provided terminology at training time if it holds true that when a terminology entry $(t_s,t_t)$ is annotated in the source, the target side $t_t$ is present in the reference. For this reason we annotate only terminology pairs that fit this criterion. 
The term bases used in the experiments are quite large and annotating all matches leads to most of the sentences containing term annotations. Since we want to model to perform equally well in a baseline, constraint-free condition, we limit the number of annotations by randomly ignoring some of the matches. 

A sentence ${\bf s}$ may contain multiple matches from a term base, but we keep the longest match in the case of overlapping source terms. Moreover, when checking for matches of a term inside a sentence, we apply \textit{approximate matching} to allow for some morphological variations in the term. In our current implementation, we use a simple character sequence match, allowing for example for base word forms to be considered matches even if they are inflected or as part of compounds.

\section{Experiments}

\subsection{Evaluation setting}
\paragraph{Parallel data and NMT architecture} 
We test our approach on the WMT 2018 English-German news translation tasks\footnote{http://www.statmt.org/wmt18/translation-task.html}, by training models on Europarl and news commentary data, for a total 2.2 million sentences. The baselines use this train data as is. For the other conditions sentences containing term annotations are added amounting to approximately 10\% of the original data. We limit the amount of data added (by randomly ignoring some of the matched terms) as we want the model to work equally well when there are no terms provided as input. Note that these sentences are from the original data pool and therefore no actual new data is introduced.

We tokenize the corpora using Moses \cite{Moses} and perform joint source and target BPE encoding \cite{Sennrich2016} to a vocabulary of 32K tokens. We use the source factor streams described in the previous section which are broadcast from word streams to BPE streams in a trivial way. We embed the three values of this additional stream into vectors of size 16 {and concatenate them to the corresponding sub-word embeddings}.
We train all models using a transformer network \cite{vaswani2017attention} with two encoding layers and two decoding layers, shared source and target  embeddings, and use the Sockeye toolkit~\cite{sockeye:2018} {(see full training configuration in the Appendix)}. {The WMT newstest 2013 development set} is used to compute the stopping criterion and all models are {trained for a minimum of 50 and a maximum of 100} epochs. 
We compare the two methods we propose, \emph{train-by-appending} and \emph{train-by-replace} with the constrained decoding algorithm of \newcite{PostV18} available in sockeye \newcite{sockeye:2018} in identical conditions and using a beam size of 5.

\paragraph{Terminology databases}
We extracted the English-German portions of two publicly available term bases, Wiktionary and IATE.\footnote{Available at \url{https://iate.europa.eu} and \url{https://www.wiktionary.org/}} In order to avoid spurious matches, we filtered out entries occurring in the top 500 most frequent English words as well as single character entries. We split the term bases into train and test lists by making sure there is no overlap on the source side.

\subsection{Results} 
We perform our evaluation on WMT newstest 2013/2017 as development (dev) and test sets respectively and use the test portions of Wiktionary and IATE to annotate the test set.\footnote{\url{https://github.com/mtresearcher/terminology_dataset}} 
We select the sentences in which the term is used in the reference and therefore the copy behaviour is justified. 
The test set extracted with the Wiktionary term base contains 727 sentences and 884 terms, while the IATE one contains 414 sentences and 452 terms. 

Table \ref{tab:cd:exactmatch:results} shows the results. 
We report decoding speed, BLEU scores, as well as term use rates, computed as the percentage of times the term translation was generated in the output out of the total number of term annotations. 

\paragraph{Term use rates and decoding speed} 
The first observation we make is that the baseline model already uses the terminology translation at a high rate of 76\%. Train-by-appending settings reach a term usage rate of around 90\%  while train-by-replace reaches even higher usage rates (93\%-94\%) indicating that completely eliminating the source term enforces the copy behaviour even more strongly. 
All these compare favourably to constrained decoding which reaches 99\% on Wiktionary but only 82\% on IATE.\footnote{We ran an additional experiment with a larger beam size of 20 and confirmed that constrained decoding can reach 99\% term use on IATE, however at a drastic latency cost.} 

Second, the decoding speed of both our settings is comparable with that of the baseline, thus three
times faster than the translation speed of 
constrained decoding (CD). This is an important difference because a three-fold increase of decoding time can hinder the use of terminology in latency-critical applications. Notice that decoding times were measured by running experiments with batch size 1 on a single GPU P3 AWS instance.\footnote{The reason for using batch size 1 is that the CD implementation does not yet offer an optimized batched version.} 

\begin{table}[ht]
\small
    \centering
    \begin{tabular}{l|lll}
    \hline
         &  \multicolumn{3}{c}{Wikt} \\
         \hline
         Model & Term\%  & BLEU ($\Delta$) & Time(s) \\
         \hline
         Baseline & 76.9 & 26.0 & 0.19  \\ 
         \hline      
         Constr. dec.  & 99.5 & 25.8 (-0.2) & 0.68 \\
         \hline
         Train-by-app. & 90.7 & 26.9 (+0.9)$\uparrow$ & 0.19 \\
         Train-by-rep. & 93.4 & 26.3 (+0.3) & 0.19 \\
        \hline
         \hline
         & \multicolumn{3}{c}{IATE } \\
         \hline
         Model & Term\%  & BLEU ($\Delta$) & Time(s) \\
         Baseline & 76.3 & 25.8 & 0.19 \\ 
         Constr. dec.  & 82.0 & 25.3 (-0.5)$\downarrow$ & 0.68 \\
         Train-by-app. & 92.9 & 26.0 (+0.2) & 0.19 \\
         Train-by-rep. & 94.5 & 26.0 (+0.2) & 0.20 \\
         \hline
    \end{tabular}
    \caption{ \small Term usage percentage and BLEU scores of systems supplied with correct term entries, exactly matching the source and the target. 
    We also provide the P99 latency numbers in seconds, i.e. the time to decode.
    $\uparrow$ and $\downarrow$ represent significantly better and worse systems than the baseline system at a p-value $<$ 0.05.}
    \label{tab:cd:exactmatch:results}
\end{table}

\begin{table*}[ht]
\small
    \centering
    \begin{tabular}{r|p{125mm}}
    \hline
       src & Plain clothes officers from Dusseldorf's police force managed to \textbf{arrest} two women and two men,  aged between 50 and 61, on Thursday. \\  
       constr dec & Plain Kleidungsbeamte der Polizei Dusseldorf konnten am Donnerstag zwei Frauen und zwei M{\"a}nner im  Alter von 50 bis 61 \textbf{Festnahme} \textbf{festzunehmen}.\\
       train-by-app & Plain Kleidungsbeamte der Polizei von Dusseldorf konnten am Donnerstag zwei Frauen und zwei M{\"a}nner \textbf{festzunehmen} , die zwischen 50 und 61 Jahre alt waren.\\
       ref & Zivilfahndern der Düsseldorfer Polizei gelang am Donnerstag die \textbf{Festnahme} von zwei Frauen und zwei Männern im Alter von 50 bis 61 Jahren.\\
       \hline
       src & The letter extends an offer to cooperate with German authorities ``when the difficulties of this \textbf{humanitarian} situation have been resolved" .\\
       constr dec &  Das Schreiben erweitert ein Angebot zur Zusammenarbeit mit den deutschen Beh{\"o}rden, ``wenn die Schwierigkeiten dieser \textbf{humanit{\"a}r} gel{\"o}st sind".\\
       train-by-app & Das Schreiben erweitert ein Angebot zur Zusammenarbeit mit den deutschen Beh{\"o}rden, ``wenn die Schwierigkeiten dieser \textbf{humanit{\"a}ren} Situation gel{\"o}st sind."\\
       ref & "In seinem Brief macht Snowden den deutschen Beh{\"o}rden ein Angebot der Zusammenarbeit, „wenn die Schwierigkeiten rund um die \textbf{humanit{\"a}re} Situation gel{\"o}st wurden“ . \\
    \hline     
    \end{tabular}
    \caption{\small Examples in which constrained decoding leads to lower translation quality due to strict enforcement of constraints. The terms are \textit{arrest} $\rightarrow$ \textit{Festnahme} and \textit{humanitarian} $\rightarrow$ \textit{humanit{\"a}r} (IATE terminology)}
    \label{tab:example_1}
\end{table*}

\begin{table}[ht]
\small
    \centering
    \begin{tabular}{l|l|l}
    \hline
         & Wiktionary & IATE  \\
         \hline
         Model & BLEU ($\Delta$) & BLEU ($\Delta$) \\
         \hline
         Baseline & 25.0 & 25.0 \\
         \hline 
         Constr. dec. & 24.1 (-0.9)$\downarrow$ & 23.7 (-1.3)$\downarrow$ \\
         \hline
         Train-by-app. & 25.0 (0.0) & 25.4 (+0.4) \\
         Train-by-rep. & 24.8 (-0.2) & 25.3 (+0.3) \\
          \hline
    \end{tabular}
    \caption{\small Machine translation results of systems supplied with term entries showing exact source matches and approximate reference matches. $\downarrow$ represent significantly worse system than baseline with a p-value $<$ 0.05.}
    \label{tab:cd:relaxedmatch:results}
\end{table}

\paragraph{Translation quality} Surprisingly, we observe significant variance w.r.t BLEU scores. Note that the terminologies affect only a small part of a sentence and most of the times the baseline already contains the desired term, therefore high BLEU variations are impossible on this test set. Constrained decoding does not lead to any changes in BLEU, other than a decrease on IATE with a small beam size of 5. However, all train-by- models show BLEU increases (+0.2 to +0.9), in particular the train-by-appending ones which have a lower terminology use rate. When examining the errors of the methods we observe cases in which constrained decoding alters the translation to accommodate a term even if a variation of that term is already in the translation as in the \textit{festzunehmen/Festnahme} example of Table \ref{tab:example_1} (and sometimes even if the identical term is already used). A closer look at previous constrained decoding literature shows that most of the evaluations are performed differently than in this paper: The data sets contain only sentences for which the reference contains the term \textit{and also the baseline fails to produce it}. This is an ideal setting which we believe to mimic few, if any, real world applications. 

We observed an additional surprisingly positive behavior with our approach which constrained decoding does not handle: in some cases, our models generate morphological variants of terminology translations provided by the term base. 
Following up on this we set up an additional experiment by extending the previous set to also include approximate matches on the target side (identical to the approximate match in training explained in Section 2.1).

Table \ref{tab:cd:relaxedmatch:results} shows these results.
We observe that this test case is already more difficult for constrained decoding as well as for train-by-replace, most likely due to the removal of the original source side content. {On the other hand, train-by-append still performs better than the baseline,} while constrained decoding shows significant BLEU score reductions of 0.9-1.3 BLEU points. The \textit{humanitarian} $\rightarrow$ \textit{humanit{\"a}r} example in Table~\ref{tab:example_1} is a representative of the errors introduced by constrained decoding in case of source matching terms whose target side needs to be inflected.

\section{Conclusion}
While most of previous work on neural MT addressed the integration of terminology with constrained decoding, we proposed a \textit{black-box} approach in which a generic neural MT architecture is directly trained to learn how to use 
an external terminology that is provided at run-time. 
We performed experiments in a \textit{zero-shot} setting, showing that the copy behaviour is triggered at test time with terms that were never seen in training. In contrast to constrained decoding, we have also observed that the method exhibits flexible use of terminology as in some cases the terms are used in their provided form while other times inflection is performed. \footnote{\citet{LuongPM15} and SYSTRAN’s Pure NMT system \citep{Systran2016} are an exception to the constrained decoding approach as they replace entities with special tags that remain unchanged during translation and are replaced in a post-processing step. However this method also lacks flexibility, as the model will always replace the placeholder with the same phrase irrespective of grammatical context. We leave comparison to their approach to future work.}

\noindent
To our knowledge there is no existing work that has a better speed vs performance trade-off than our method in the space of constrained decoding algorithms for neural MT, which we believe makes it particularly suitable for production environments. 

\section{Aknowledgments}
The authors would like to thank Wael Hamza, Faisal Ladhak, Mona Diab and the anonymous reviewers for their advice and comments.

\bibliographystyle{acl_natbib}
\bibliography{naaclhlt2019}

  \clearpage
  \begin{appendices}



\section*{Appendix A}
\textbf{NMT Sockeye train parameters}

\begin{verbatim}
encoder-config:
  act_type: relu
  attention_heads: 8
  conv_config: null
  dropout_act: 0.1
  dropout_attention: 0.1
  dropout_prepost: 0.1
  dtype: float32
  feed_forward_num_hidden: 2048
  lhuc: false
  max_seq_len_source: 101
  max_seq_len_target: 101
  model_size: 512
  num_layers: 2
  positional_embedding_type: fixed
  postprocess_sequence: dr
  preprocess_sequence: n
  use_lhuc: false

decoder config: 
  act_type: relu
  attention_heads: 8
  conv_config: null
  dropout_act: 0.1
  dropout_attention: 0.1
  dropout_prepost: 0.1
  dtype: float32
  feed_forward_num_hidden: 2048
  max_seq_len_source: 101
  max_seq_len_target: 101
  model_size: 512
  num_layers: 2
  positional_embedding_type: fixed
  postprocess_sequence: dr
  preprocess_sequence: n
 
config_loss: !LossConfig
  label_smoothing: 0.1
  name: cross-entropy
  normalization_type: valid
  vocab_size: 32302
  
config_embed_source: !EmbeddingConfig
  dropout: 0.0
  dtype: float32
  factor_configs: null
  num_embed: 512
  num_factors: 1
  vocab_size: 32302  

config_embed_target: !EmbeddingConfig
  dropout: 0.0
  dtype: float32
  factor_configs: null
  num_embed: 512
  num_factors: 1
  vocab_size: 32302
\end{verbatim}

  \end{appendices}

\end{document}